\documentclass[conference]{IEEEtran}
\IEEEoverridecommandlockouts
\usepackage{cite}
\usepackage{amsmath,amssymb,amsfonts}
\usepackage{algorithmic}
\usepackage{graphicx}
\usepackage{textcomp}
\usepackage{xcolor}
\usepackage{float} 
\usepackage{url}
\usepackage{fancyhdr}

\newcommand{\linebreakand}{%
  \end{@IEEEauthorhalign}
  \hfill\mbox{}\par
  \mbox{}\hfill\begin{@IEEEauthorhalign}
}
\makeatother

\def\BibTeX{{\rm B\kern-.05em{\sc i\kern-.025em b}\kern-.08em
    T\kern-.1667em\lower.7ex\hbox{E}\kern-.125emX}}
\begin{document}

\setlength{\headheight}{20pt}

\fancypagestyle{firstpage}{
    \fancyhf{}
    
    \fancyhead[L]{\small \textcolor{red} {This paper has been accepted in 10th International Conference on Signal processing and Communication (IEEE ICSC 2025). Published version of the paper will be available in IEEE Xplore very soon.}}
    \fancyhead[C]{}
    \fancyhead[R]{}
    \renewcommand{\headrulewidth}{0pt}
    \renewcommand{\footrulewidth}{0pt}
}

\title{Automated Attendee Recognition System for Large-Scale Social Events or Conference Gathering}

\author{
    \IEEEauthorblockN{1\textsuperscript{st} Dhruv Motwani}
    \IEEEauthorblockA{\textit{Head of GenAI} \\
    \textit{Avahi}\\
    San Francisco, CA, USA \\
    dhruv.motwani@avahitech.com}
    \and
    \IEEEauthorblockN{2\textsuperscript{nd} Ankush Tyagi}
    \IEEEauthorblockA{\textit{Software Development Manager} \\
    \textit{Ericsson}\\
    Austin, Texas, USA \\
    ankush.tyagi@ericsson.com}
    \and
    \IEEEauthorblockN{3\textsuperscript{rd} Vipul Dabhi }
    \IEEEauthorblockA{\textit{Department of Information Technology} \\
    \textit{Dharmsinh Desai University}\\
    Nadiad, India \\
    vipuldabhi.it@ddu.ac.in}
    \and
    \IEEEauthorblockN{4\textsuperscript{th} Harshadkumar Prajapati }
    \IEEEauthorblockA{\textit{Department of Information Technology} \\
    \textit{Dharmsinh Desai University}\\
    Nadiad, India \\
    prajapatihb.it@ddu.ac.in}
}

\maketitle

\thispagestyle{firstpage}

\begin{abstract}
Manual attendance tracking at large-scale events, such as marriage functions or conferences, is often inefficient and prone to human error. To address this challenge, we propose an automated, cloud-based attendance tracking system that uses cameras mounted at the entrance and exit gates. The mounted cameras continuously capture video and send the video data to cloud services to perform real-time face detection and recognition. Unlike existing solutions, our system accurately identifies attendees even when they are not looking directly at the camera, allowing natural movements, such as looking around or talking while walking. To the best of our knowledge, this is the first system to achieve high recognition rates under such dynamic conditions. Our system demonstrates overall 90\% accuracy, with each video frame processed in 5 seconds, ensuring real-time operation without frame loss. In addition, notifications are sent promptly to security personnel within the same latency. This system achieves 100\% accuracy for individuals without facial obstructions and successfully recognizes all attendees appearing within the camera’s field of view, providing a robust solution for attendee recognition in large-scale social events.
\end{abstract}

\begin{IEEEkeywords}
attendance tracking, face recognition, face detection, cloud-based system
\end{IEEEkeywords}

\section{Introduction}
\label{introduction}
In today's fast-paced world, ensuring efficient management of large-scale events such as conferences, weddings, and social gatherings has become increasingly important. One critical task is recording attendance \cite{ishaq2023iot, nuhi2020smart, islam2017development, mohamed2012fingerprint, hsiung2011performance, raj2020face}, which often involves manual efforts, leading to time consumption, errors, and inefficiencies. With advancements in camera technology and artificial intelligence, it has become feasible to automate attendance tracking \cite{raj2020face, kawaguchi2005face, damale2018face, mery2019student}. This can significantly streamline operations, providing real-time monitoring and minimizing the need for human intervention. Automatic attendance systems \cite{nuhi2020smart, islam2017development, mohamed2012fingerprint, hsiung2011performance, raj2020face} can also improve the accuracy and reliability of attendance records.

Despite advancements in AI-based monitoring systems \cite{raj2020face, kawaguchi2005face, damale2018face}, the challenge of automating attendance through mounted cameras in venues has not been fully addressed. Existing systems often rely on manual interaction or less sophisticated technologies, which are not well-suited for large gatherings in open environments like marriage halls or event venues. Issues such as crowd density, lighting conditions, and varying camera angles further complicate the problem \cite{mery2019student}. Thus, a robust, scalable solution is required to ensure accurate identification and attendance tracking in these dynamic environments.

Currently, several attendance tracking systems use biometric devices, RFID (Radio Frequency Identification) tags \cite{ishaq2023iot, ansari2011automation}, or mobile applications \cite{nuhi2020smart} to log attendees' presence. However, these approaches either require active participation from users or are limited by the need for specialized hardware. While facial recognition technology \cite{raj2020face} has been applied in controlled environments such as classrooms \cite{kawaguchi2005face}, its adoption in crowded and unregulated environments like function halls is still in its infancy. Major limitations include performance degradation in varying lighting conditions, difficulty in recognizing faces within dense crowds, and concerns regarding privacy and security.

This paper proposes an automated, camera-based attendance system architecture using advanced facial recognition algorithms and cloud technology that can accurately identify individuals in a wide range of environments, from well-lit conference venues to dimly lit wedding halls. The system leverages machine learning techniques to adapt to different lighting conditions, crowd densities, and angles of observation, offering a non-intrusive, hands-free solution for event organizers. The novelty of this approach is in its ability to function in challenging, real-world conditions where existing systems can not perform well. Our proposed system enhances both accuracy and convenience, making attendance tracking seamless.

The remainder of this paper is organized as follows: Section \ref{related-work} provides an overview of the related work and existing systems in automated attendance tracking. Section \ref{proposed-work} discusses the proposed system, including the hardware setup, architecture design, and implementation details. Section \ref{experiments-results} presents the experimental setup, data collection, evaluation metrics, and results. Finally, Section \ref{conclusion} concludes the paper with a summary of findings and directions for future research.

\section{Background Theory and Related Work}
\label{related-work}
Traditional attendance tracking systems rely on manual sign-ins or the use of RFID cards \cite{ishaq2023iot, ansari2011automation}, QR codes \cite{nuhi2020smart}, and mobile apps \cite{islam2017development}. These methods, while useful, often require user participation, which leads to inefficiencies in large gatherings. Some researchers have explored automated attendance systems using biometric data such as fingerprints \cite{mohamed2012fingerprint} and iris scans \cite{hsiung2011performance}. Such systems require close-range interaction with hardware, making them impractical for large event spaces like marriage halls. There are different mechanisms using which attendance system can be built. A recent survey work in \cite{ali2022automated} presents a concise survey on automated attendance system. Their work analyzed 90 research papers and presented their findings concisely, which help beginners a lot to get quick views and findings in the domain of Automated attendance management systems.

Facial recognition technology \cite{raj2020face} has gained significant attention in biometric-based attendance systems. Early works by \cite{kawaguchi2005face} demonstrated the efficacy of facial recognition algorithms for low-scale environments such as classrooms. More recent studies have integrated machine learning to improve recognition rates under controlled conditions \cite{damale2018face}  A recent work in \cite{thaleeparambil2024integrated} presents RFID based system built using Open Source hardware modules such as ESP-32 and Arduino. Another recent work in \cite{venkatakrishnan2024design} also uses hardware module such as ESP-32 and ESP32-CAM hardware components for capturing students' attendance to avoid proxies attempted by students. These various works focus on a closed and controlled environment; however, challenges remain when these systems are deployed in crowded, unregulated spaces \cite{mery2019student}.

Work by Harikrishnan et al. in \cite{harikrishnan2019vision} has presented a real-time attendance monitoring system using deep learning; however, the system needs to improve accuracy. One major challenge facial recognition systems face in real-world applications is the impact of lighting and crowd density on performance. Various systems tested in outdoor environments have shown a marked decline in accuracy. A survey work by Adjabi et al. \cite{adjabi2020past} concisely presents the face recognition methods of past, present, and future, which can be very useful for newbies in the field. 

To the best of our knowledge, automating attendance tracking at large-scale events such as weddings or conferences has not been attempted or reported in the literature. The use of mobile apps can track entry and exit, but these methods still require active user participation. Another possibility is to utilize drones to monitor crowds. While significant progress has been made in automating attendance systems, there is still a gap in the literature regarding the use of facial recognition for large-scale, real-time attendance tracking at events with varying environmental conditions. If a system capable of handling crowded environments can be implemented, it can be used for multiple purposes. In this paper, we propose a solution that is robust and can handle dynamic lighting and high traffic density without compromising accuracy.

\section{Proposed Work}
\label{proposed-work}
\subsection{Proposed System}\label{AA}
Presently, several methods are used to address the problem of invitation validation at events. The most common solution involves employing staff members or private security personnel who manually verify attendees. This approach, while straightforward, often leads to inefficiencies, as attendees are required to wait in queues, and manual validation can be slow and error-prone.

Another widely adopted technique is the use of bar codes, QR codes \cite{nuhi2020smart}, or RFID tags attached to the invitations or provided to attendees. These technologies allow for quicker validation through scanning devices. However, they still require individuals to carry a physical or digital item (such as a card or mobile device) to gain entry. Moreover, this process still often involves standing in lines and can be cumbersome for high-profile guests or those attending large-scale events.

Both of these approaches, while functional, impose certain limitations. They require attendees to carry specific items and endure potentially lengthy wait times, which diminishes the overall event experience. As a result, these solutions are not entirely intuitive or seamless, particularly for high-profile events where efficiency and attendee convenience are paramount.

\begin{figure*}[t!]
  \centering
    \includegraphics[width=0.85\textwidth]{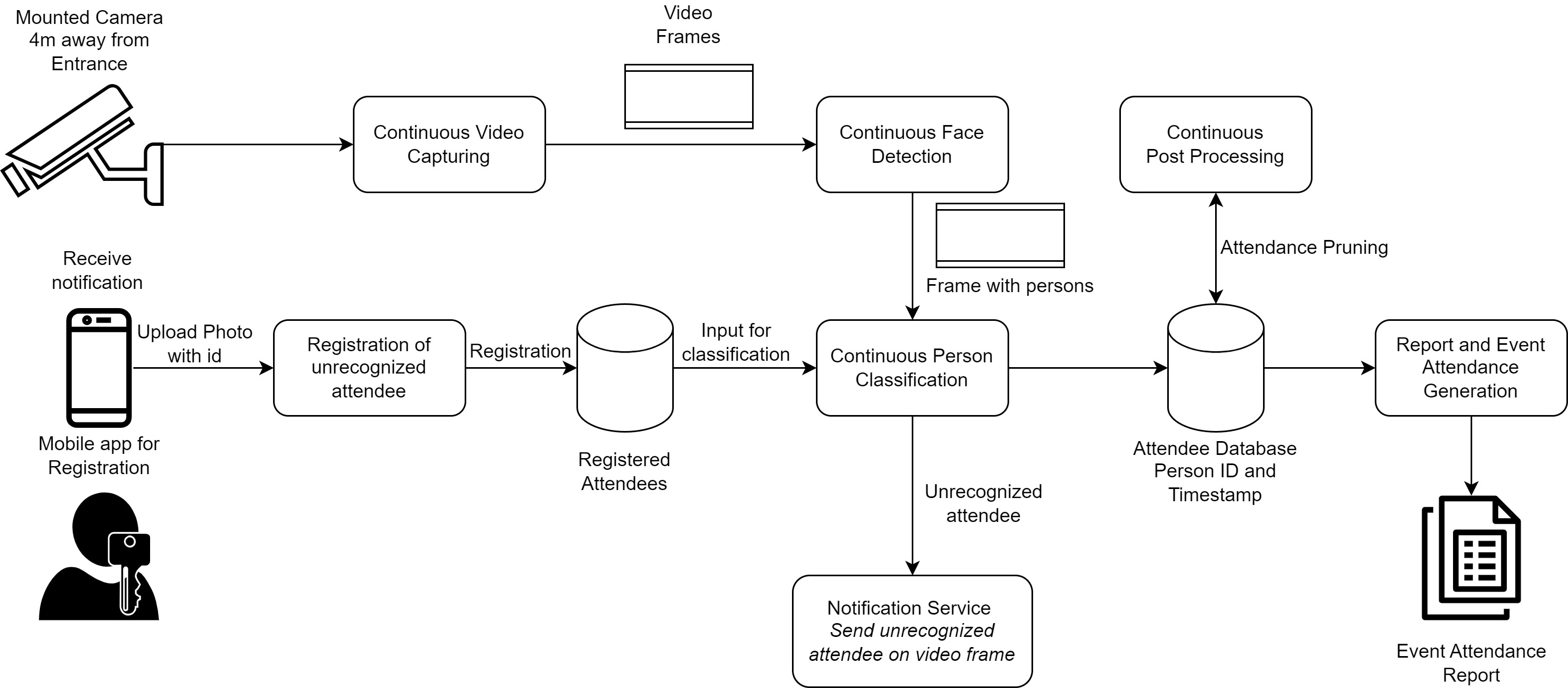}
  \caption{High-level Architecture of the Proposed System: Automated Attendee Recognition System}
  \label{fig:proposed-system}
\end{figure*}

We propose a solution in Figure \ref{fig:proposed-system} that leverages advanced computer vision technology to streamline the invitation validation process. A camera system is installed 4 metres away from the main entrance of the event venue. As individuals approach, the camera captures their images and uses facial recognition algorithms to validate their identities against a pre-registered database of invited attendees.

Once a person’s identity is confirmed, they are seamlessly granted entry without needing to present any physical items or wait in long queues. If an individual is not pre-registered or cannot be verified by the system, the event official stationed at the entrance will be notified, with the person’s image displayed for reference. This allows the official to quickly assess the situation and proceed with manual validation or initiate the registration process if required.
This approach not only reduces waiting times but also provides a more intuitive and frictionless experience for attendees. By eliminating the need to carry invitation materials, the proposed solution enhances security, improves efficiency, and delivers a smoother validation process, especially for high-profile events.

\subsection{System Setup}
The proposed setup requires a high-quality camera for effective image capture and facial recognition. In our current implementation, we have successfully tested the system with a 2MP HikVision camera. However, the number of cameras required will depend on the attendees' volume and the walkway's width to ensure adequate coverage and uninterrupted flow.

The system requires a walkway of approximately 4-5 meters for optimal performance. This distance provides sufficient time for the camera to capture clear images and for the facial recognition model to detect and process attendees in real-time, ensuring smooth validation without delays.

The entire system is designed to be cloud-based, eliminating the need for local database storage or on-site system management. All image recognition, processing, and attendee validation occur in the cloud, which ensures scalability and reduces the infrastructure burden for event organizers.



\subsection{Implementation of the Proposed System}
\begin{figure*}[t!]
  \centering
    \includegraphics[width=0.85\textwidth]{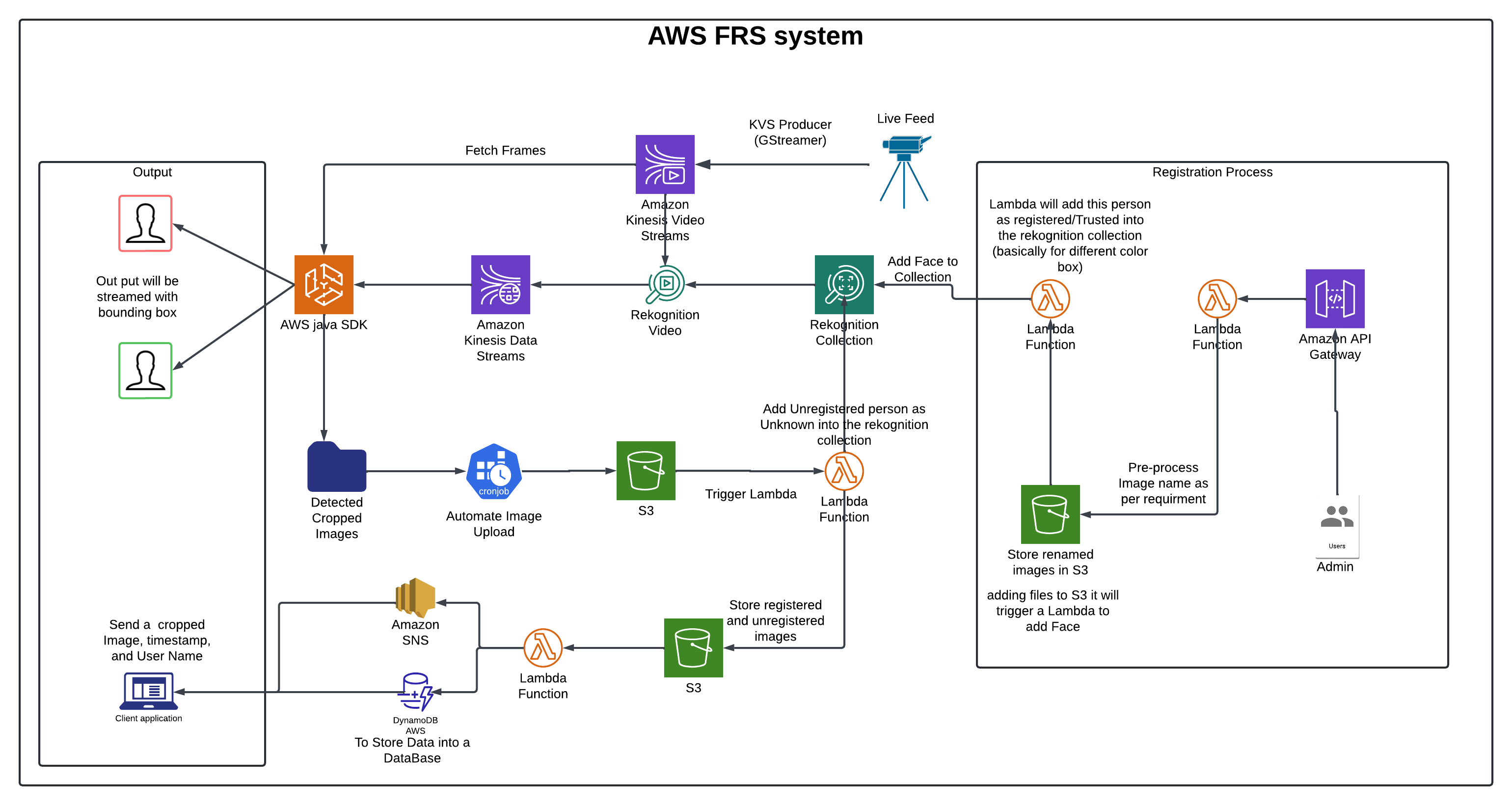}
  \caption{Implementation of the proposed system: AWS Cloud components and Lambda Functions for processing}
  \label{fig:implementation-diagram}
\end{figure*}

Figure \ref{fig:implementation-diagram} illustrates the entire workflow of the system, from capturing images to delivering output via G-Streamer, with the help of a Java Parser library. We discuss the main components.

\begin{itemize}
    \item Image Capture: Cameras (e.g., 2MP HikVision cameras) are strategically placed to capture images of individuals approaching the event entrance. The number of cameras is adjusted according to the expected crowd size and the width of the walkway.
    
    \item Java Parser (Local Processing): For security tracking purposes, each camera is paired with a local instance of a Java Parser. This parser processes the captured images offline to track individuals in real-time, ensuring swift communication between the camera and the cloud. The parser leverages high-performance CPUs to manage the processing and efficient data transfer to the cloud.

    \item Cloud-Based Processing: Once images are captured, they are sent to the cloud, where server-less components handle the main image processing. The most crucial element in this architecture is AWS Rekognition, which performs facial recognition by comparing captured images with a pre-registered attendee database.

    \item G-Streamer Output: After processing, the output is displayed using G-Streamer, which provides real-time feedback, indicating whether the individual is validated or requires manual verification.

    \item Server-less Components: Several server-less components orchestrate the workflow, ensuring scalability, fault tolerance, and minimal maintenance. These components manage the communication between the camera system, Java Parser, AWS Rekognition, and G-Streamer, making the entire solution seamless and easy to deploy.

    \item Hybrid Cloud-Local System: While most components operate on the cloud, the Java Parser runs locally to ensure real-time tracking without burdening the cloud infrastructure. This hybrid model optimizes both security and performance, providing a seamless attendee validation experience.

\end{itemize}

\section{Experiments and Results}
\label{experiments-results}
\subsection{Dataset Preparation}
A dataset of 100 people with variety and similarity in their faces was created. The dataset included 39 females and 61 male persons. Major characteristics of the faces of all participants are detailed in Table \ref{tabular-dataset}. We considered various face features that are generally found to test the accuracy of the proposed system. Out of 100 people, 65 people were available for capturing their face images. We prepared a dataset by capturing three images of one person: (1) front face, (2) face at an angle, (3) side face. For the remaining 35 people, we captured their facial images by scanning their face images from Aadhaar cards, school identity cards, and PAN cards.

\begin{table}[!ht]
\begin{center}
\caption{Details of Participants used in Preparation of the Dataset}
\label{tabular-dataset}
\begin{tabular}{p{2in}r}
\hline
Participant Information & Count \\
\hline
Number of participants  &    100\\
\hline
Female participants  &    39\\
Male participants  &  61  \\
\hline
Boys participants (Aged 10 to 18 )  &    7\\
Male participants (Aged 19 to 60)  &   48 \\
Male participants (Aged above 60)  &    6\\
\hline
Male participants with a beard  &  4  \\
Male participants with a cap  &  1  \\
Male participants with Spectacles  &   6 \\
Male participants with a mustache  &  6  \\
\hline
Girls participants (Aged 10 to 18)  &    4\\
Female participants (Aged 19 to 60)  &   33 \\
Female participants (Aged above 60)  &   2 \\
\hline
Female participants with Spectacles  &  3  \\
\hline
Female participants with long hair  &  22  \\
Female participants with short hair  &  17  \\
\hline
\end{tabular}
\end{center}
\end{table}

\begin{figure*}[t!]
  \centering
    \includegraphics[width=0.85\textwidth]{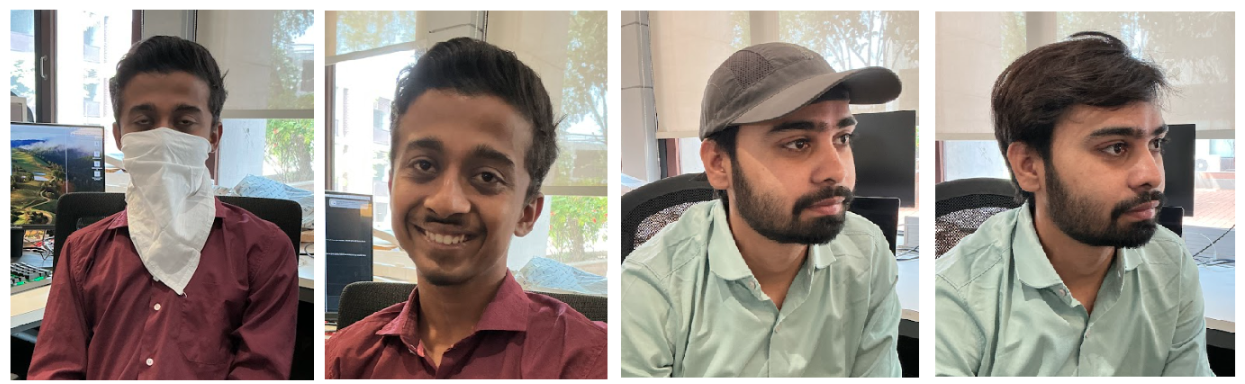}
  \caption{Sample images of dataset preparation of available participants}
  \label{fig:sample-faces}
\end{figure*}

\begin{figure}[t!]
  \centering
    \includegraphics[width=0.35\textwidth]{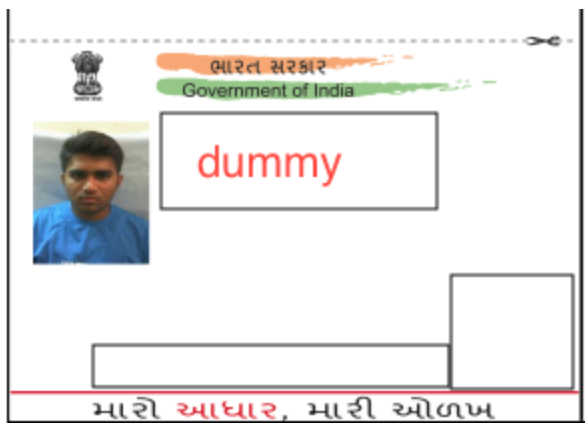}
  \caption{A sample image of dataset preparation of unavailable participants}
  \label{fig:aadhaar-card}
\end{figure}

While preparing the dataset, all face images were converted into size 400 X 300 pixels. Few images of dataset preparation are shown in Figure \ref{fig:sample-faces}. For all available participants, images were captured using the same camera. For unavailable participants, we extracted their face images from their Aadhaar cards or PAN cards. Figure \ref{fig:aadhaar-card} shows a sample Aadhaar card showing face image of a person.

\subsection{Experiments Settings}
We conducted various experiments to assess the performance, accuracy, and reliability of the proposed automated attendee recognition system. The experiments simulated different real-world scenarios to evaluate how effectively the system handles attendee detection and recognition under varying conditions.
\begin{itemize}
  \item A single person in the video
  \item Multiple persons in the video all at same distance
  \item Multiple persons in the video: some are close to the camera and some are far
  \item Persons having a similar overall look (having a beard, wearing glasses, having a mustache, wearing masks)
\end{itemize}

\emph{Input Data}
The experiments used live video feeds capturing attendees as they entered and exited the venue. These video frames contained both attendees registered in the database and individuals who were not, simulating a realistic setting with mixed populations.

\emph{Output} includes the following:
\begin{itemize}
  \item Detecting Entry Time: The system recorded the precise moment when an attendee made their entry into the event area by identifying them in the live video stream.
  \item Detecting Exit Time: Similarly, the system logged the exact time when the individual exited the venue from exit gate.
\end{itemize}

\emph{Performance Measures} includes the following:
\begin{itemize}
  \item Accuracy: The accuracy of detecting the identity of each attendee was measured across different experimental setups. Special attention was given to how well the system performed in cases with individuals having similar appearances or wearing masks.
  \item Latency: We measured the time taken by the system to detect and verify an attendee’s identity from the moment they appear in a video frame. Average latency was maintained at approximately 5 seconds, which includes time to process a frame only 1 second, which was a critical factor in ensuring real-time processing for large-scale events. 
  \item Notification Time: Once a person’s identity was detected, a notification was promptly sent to security personnel at the entry gate. The time taken to send this notification was recorded to ensure timely alerts, and the system successfully maintained this time within a 10-second window.  
\end{itemize}

\subsection{Results and Discussion}
\begin{figure*}[t!]
  \centering
    \includegraphics[width=0.65\textwidth]{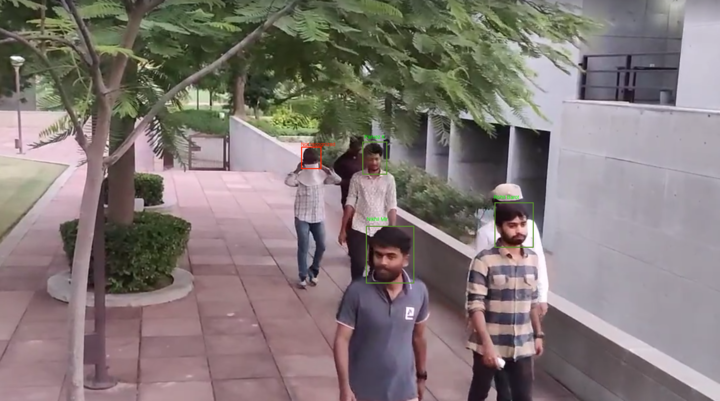}
  \caption{Testing of the proposed system: green bounding boxes indicate persons detected and recognized successfully (3 persons in this frame) and red bounding boxes indicate a face is detected, but the person is not recognized (1 person in this frame)}
  \label{fig:testing-people}
\end{figure*}

To evaluate the proposed system, we set up camera mounting in our office campus as per the proposed architecture. We performed multiple experiments during different times of the day. There was a gap of 10 days between participant registration and evaluation of the system. People walked on a pathway as if they are unaware that cameras were mounted. Persons were going through mounted cameras as per different experiment settings: single person, group of persons, persons walking behind some other persons, a person wearing mask, a person wearing cap, etc. A sample video frame with person recognition is shown in Figure \ref{fig:testing-people}. For each video frame, the system writes the person’s name on top of the bounding box. The system also records the timestamp and the name and id of the person and stores it in the database, as shown in implementation diagram in Figure \ref{fig:implementation-diagram}.

The average performance of multiple experiments is shown in Table \ref{table-perf}.

\begin{table*}[!ht]
\begin{center}
\caption{Performance of the Proposed System on Different Performance Measures}
\label{table-perf}
\begin{tabular}{p{4in}p{2in}}
\hline
Performance Measure & Performance Value \\
\hline
Accuracy of detecting the identity of an attendee  &   97\% \\
Accuracy of detecting the identity of an attendee wearing a cap    &   1 out of 1 \\
Accuracy of detecting a single person in a video frame  &   100\% \\
Accuracy of detecting multiple persons in a video frame  &   90.34\% \\
Accuracy of detecting persons with side faces in a video frame  &   90.00\% with less than 60 degree \\
Latency of detecting the identity of an attendee  &   within 5 seconds, with frame processing  within only 1 second \\
Time to send a notification to the security person  &   within 10 seconds \\
\hline

\end{tabular}
\end{center}
\end{table*}

For the pilot study, we recognized persons in each video frame and recorded the timestamp of recognition and the name and ID of a person in the result dataset. We decided to recognize persons in each video frame because while persons walk they may not keep their faces straight staring at a fixed point. They may look around or talk with other accompanying persons while walking. Therefore, there were multiple entries of a person in the database due to a person recognized in multiple video frames. However, to use this system, in a real scenario, we wanted to have a single entry for one person appearing in an entire clip of around 6 to 8 seconds, during which a person remains in focus of the mounted camera. To handle these multiple entries, we have implemented a pruning algorithm, which removes duplicate entries by moving a moving window of 10 seconds on recognition result, which is time-ordered rows containing a timestamp, person name, and ID.

While the proposed system successfully automates attendee recognition with high accuracy, there are several directions for further improvement. For future work, we plan to optimize various cloud resources used for storing video frames, and processing continuous video frames. We would perform experiments on large population size with focus on enhancing the system’s performance for individuals with facial obstructions, such as masks, hats, or sunglasses, which were not fully addressed in the current implementation. Further testing across diverse environments and lighting conditions is necessary to ensure robustness in various real-world scenarios. These improvements will broaden the applicability of the system to a wider range of events and settings.

\section{Conclusion}
\label{conclusion}
In this paper, we proposed an automated attendee recognition system for large-scale social events or conferences and implemented this system using cloud services. The system was evaluated using different experiment scenarios. The results indicate that our proposed automated attendee recognition system provides an efficient and accurate solution for tracking attendees at large-scale social events, such as marriage functions and conferences. By utilizing continuous video capture and real-time face detection, the system can accurately recognize individuals even when they are engaged in natural movements like talking or looking around. With an overall accuracy rate of  90\% and latency as low as 5 seconds per frame, our implemented system demonstrates its capability to function effectively in dynamic environments. Additionally, the prompt notification feature further enhances security, ensuring timely alerts to personnel. This system's ability to achieve 100\% recognition accuracy for individuals without facial obstructions highlights its robustness, making it a valuable tool for large-scale event management. Future work could focus on improving accuracy for individuals wearing accessories like masks or hats and further reducing processing time to support even larger crowds.

\bibliographystyle{IEEEtran}
\bibliography{main}

\begin{thebibliography}{10}
\providecommand{\url}[1]{#1}
\csname url@samestyle\endcsname
\providecommand{\newblock}{\relax}
\providecommand{\bibinfo}[2]{#2}
\providecommand{\BIBentrySTDinterwordspacing}{\spaceskip=0pt\relax}
\providecommand{\BIBentryALTinterwordstretchfactor}{4}
\providecommand{\BIBentryALTinterwordspacing}{\spaceskip=\fontdimen2\font plus
\BIBentryALTinterwordstretchfactor\fontdimen3\font minus \fontdimen4\font\relax}
\providecommand{\BIBforeignlanguage}[2]{{%
\expandafter\ifx\csname l@#1\endcsname\relax
\typeout{** WARNING: IEEEtran.bst: No hyphenation pattern has been}%
\typeout{** loaded for the language `#1'. Using the pattern for}%
\typeout{** the default language instead.}%
\else
\language=\csname l@#1\endcsname
\fi
#2}}
\providecommand{\BIBdecl}{\relax}
\BIBdecl

\bibitem{ishaq2023iot}
K.~Ishaq and S.~Bibi, ``Iot based smart attendance system using rfid: A systematic literature review,'' \emph{arXiv preprint arXiv:2308.02591}, 2023.

\bibitem{nuhi2020smart}
A.~Nuhi, A.~Memeti, F.~Imeri, and B.~Cico, ``Smart attendance system using qr code,'' in \emph{2020 9th mediterranean conference on embedded computing (MECO)}.\hskip 1em plus 0.5em minus 0.4em\relax IEEE, 2020, pp. 1--4.

\bibitem{islam2017development}
M.~M. Islam, M.~K. Hasan, M.~M. Billah, and M.~M. Uddin, ``Development of smartphone-based student attendance system,'' in \emph{2017 IEEE Region 10 Humanitarian Technology Conference (R10-HTC)}.\hskip 1em plus 0.5em minus 0.4em\relax IEEE, 2017, pp. 230--233.

\bibitem{mohamed2012fingerprint}
B.~K. Mohamed and C.~Raghu, ``Fingerprint attendance system for classroom needs,'' in \emph{2012 Annual IEEE India Conference (INDICON)}.\hskip 1em plus 0.5em minus 0.4em\relax IEEE, 2012, pp. 433--438.

\bibitem{hsiung2011performance}
T.~W. Hsiung and S.~S. Mohamed, ``Performance of iris recognition using low resolution iris image for attendance monitoring,'' in \emph{2011 IEEE International Conference on Computer Applications and Industrial Electronics (ICCAIE)}.\hskip 1em plus 0.5em minus 0.4em\relax IEEE, 2011, pp. 612--617.

\bibitem{raj2020face}
A.~A. Raj, M.~Shoheb, K.~Arvind, and K.~Chethan, ``Face recognition based smart attendance system,'' in \emph{2020 International conference on intelligent engineering and management (ICIEM)}.\hskip 1em plus 0.5em minus 0.4em\relax IEEE, 2020, pp. 354--357.

\bibitem{kawaguchi2005face}
Y.~Kawaguchi, T.~Shoji, W.~Lin, K.~Kakusho, M.~Minoh \emph{et~al.}, ``Face recognition-based lecture attendance system,'' in \emph{The 3rd AEARU workshop on network education}.\hskip 1em plus 0.5em minus 0.4em\relax Citeseer, 2005, pp. 70--75.

\bibitem{damale2018face}
R.~C. Damale and B.~V. Pathak, ``Face recognition based attendance system using machine learning algorithms,'' in \emph{2018 Second International Conference on Intelligent Computing and Control Systems (ICICCS)}.\hskip 1em plus 0.5em minus 0.4em\relax IEEE, 2018, pp. 414--419.

\bibitem{mery2019student}
D.~Mery, I.~Mackenney, and E.~Villalobos, ``Student attendance system in crowded classrooms using a smartphone camera,'' in \emph{2019 IEEE Winter Conference on Applications of Computer Vision (WACV)}.\hskip 1em plus 0.5em minus 0.4em\relax IEEE, 2019, pp. 857--866.

\bibitem{ansari2011automation}
A.~N. Ansari, A.~Navada, S.~Agarwal, S.~Patil, and B.~A. Sonkamble, ``Automation of attendance system using rfid, biometrics, gsm modem with. net framework,'' in \emph{2011 International Conference on Multimedia Technology}.\hskip 1em plus 0.5em minus 0.4em\relax IEEE, 2011, pp. 2976--2979.

\bibitem{ali2022automated}
N.~S. Ali, A.~H. Alhilali, H.~D. Rjeib, H.~Alsharqi, and B.~Al-Sadawi, ``Automated attendance management systems: systematic literature review,'' \emph{International Journal of Technology Enhanced Learning}, vol.~14, no.~1, pp. 37--65, 2022.

\bibitem{thaleeparambil2024integrated}
N.~Thaleeparambil, A.~Biju, and B.~Prathap, ``Integrated automated attendance system with rfid, wi-fi, and visual recognition technology for enhanced classroom security and precise monitoring,'' in \emph{2024 IEEE International Conference on Contemporary Computing and Communications (InC4)}, vol.~1.\hskip 1em plus 0.5em minus 0.4em\relax IEEE, 2024, pp. 1--6.

\bibitem{venkatakrishnan2024design}
G.~Venkatakrishnan, R.~Rengaraj, R.~Jeya, S.~Manikandan \emph{et~al.}, ``Design and implementation of automated attendance system using contactless facial recognition,'' in \emph{2024 International Conference on Power, Energy, Control and Transmission Systems (ICPECTS)}.\hskip 1em plus 0.5em minus 0.4em\relax IEEE, 2024, pp. 1--6.

\bibitem{harikrishnan2019vision}
J.~Harikrishnan, A.~Sudarsan, A.~Sadashiv, and R.~A. Ajai, ``Vision-face recognition attendance monitoring system for surveillance using deep learning technology and computer vision,'' in \emph{2019 international conference on vision towards emerging trends in communication and networking (ViTECoN)}.\hskip 1em plus 0.5em minus 0.4em\relax IEEE, 2019, pp. 1--5.

\bibitem{adjabi2020past}
I.~Adjabi, A.~Ouahabi, A.~Benzaoui, and A.~Taleb-Ahmed, ``Past, present, and future of face recognition: A review,'' \emph{Electronics}, vol.~9, no.~8, p. 1188, 2020.

\end{thebibliography}

\end{document}